\documentclass[10pt,twocolumn,letterpaper]{article}

\usepackage[pagenumbers]{wacv} 

\usepackage[table]{xcolor}
\usepackage{multirow} 
\definecolor{wacvblue}{rgb}{0.21,0.49,0.74}
\usepackage[pagebackref,breaklinks,colorlinks,allcolors=wacvblue]{hyperref}

\def\wacvPaperID{1} 
\def\confName{WACV}
\def\confYear{2026}

\title{K-Track: Kalman-Enhanced Tracking\\for Accelerating Deep Point Trackers on Edge Devices}
\author{Bishoy Galoaa, Pau Closas, Sarah Ostadabbas\\
Northeastern University\\
Boston, MA, USA\\
{\tt\small \{galoaa.b, p.closas, ostadabbas\}@northeastern.edu}
}
\begin{document}
\maketitle
\newcommand{\comparisontable}{
\begin{table}[t]
\setlength{\tabcolsep}{3pt}
\centering
\caption{Performance degradation of state-of-the-art efficient trackers on edge devices. FPS measured at 480p resolution. Color intensity indicates performance: \textcolor{green!70}{green} (real-time), \textcolor{orange!70}{orange} (marginal), \textcolor{red!70}{red} (unusable). K-Track with N=10 runs deep learning inference every 10 frames, using Kalman filtering for intermediate predictions.}
\label{tab:deployment_gap}
\footnotesize
\begin{tabular}{@{}lcc@{}}
\toprule
\textbf{Model} & \textbf{RTX Titan} & \textbf{Jetson Orin} \\
& \textbf{24GB, 130 TF} & \textbf{Nano 8GB, 40 TOPS} \\
\midrule
CoTracker3~\cite{cotracker3}(Online) & \cellcolor{green!30}15--20 & \cellcolor{orange!40}1.5--2.0 \\
CoTracker3~\cite{cotracker3}(Offline) & \cellcolor{green!30}18--25 & \cellcolor{orange!40}1.8--2.5 \\
Track-On~\cite{aydemir2025track}& \cellcolor{orange!50}8--10 & \cellcolor{red!50}0.5--0.7 \\
SpatialTracker~\cite{xiao2024spatialtracker}& \cellcolor{red!30}0.12 & \cellcolor{red!70}$<$0.1 \\
\textbf{K-Track (N=10)} & \cellcolor{green!50}\textbf{45--60} & \cellcolor{green!40}\textbf{4.5--6.0} \\
\bottomrule
\end{tabular}
\end{table}
}
\newcommand{\accuracytable}{
\begin{table}[t]
\centering
\caption{Tracking accuracy on TAP-Vid-DAVIS across keyframe intervals. Values show PCK@5px (top) and percentage retention (bottom) for the best grid size at each $N$. Best grid sizes: CoTracker3 uses Grid 20 (best baseline), SpatialTracker uses Grid 32 (best baseline), Track-On uses Grid 15 (best baseline). Detailed grid analysis in Tables 5--6. Color coding: \textcolor{green!60!black}{$>$90\%}, \textcolor{yellow!80!black}{85--90\%}, \textcolor{orange!70}{80--85\%}, \textcolor{red!70}{$<$80\%}.}
\label{tab:accuracy}
\scriptsize
\setlength{\tabcolsep}{3pt}
\begin{tabular}{@{}lc|ccccc@{}}
\toprule
\textbf{Tracker} & \textbf{Baseline} & \multicolumn{5}{c}{\textbf{Keyframe Interval } $N$} \\
& \textbf{PCK@5px} & \textbf{2} & \textbf{3} & \textbf{5} & \textbf{10} & \textbf{15} \\
\midrule
CoTracker3~\cite{cotracker3} & 94.4\% & \cellcolor{green!30}\small 91.2\% & \cellcolor{green!30}\small 79.9\% & \cellcolor{green!20}\small 74.9\% & \cellcolor{yellow!40}\small 67.2\% & \cellcolor{orange!40}\small 62.1\% \\
& (Grid 20) & \small (97\%) & \small (85\%) & \small (79\%) & \small (71\%) & \small (66\%) \\
\midrule
SpatialTracker~\cite{xiao2024spatialtracker} & 87.9\% & \cellcolor{green!30}\small 84.5\% & \cellcolor{green!30}\small 77.0\% & \cellcolor{green!20}\small 72.0\% & \cellcolor{yellow!40}\small 68.5\% & \cellcolor{orange!40}\small 61.2\% \\
& (Grid 32) & \small (96\%) & \small (88\%) & \small (82\%) & \small (78\%) & \small (70\%) \\
\midrule
Track-On~\cite{aydemir2025track} & 96.4\% & \cellcolor{green!30}\small 87.5\% & \cellcolor{green!30}\small 83.9\% & \cellcolor{green!20}\small 79.0\% & \cellcolor{yellow!40}\small 69.1\% & \cellcolor{orange!40}\small 64.0\% \\
& (Grid 15) & \small (91\%) & \small (87\%) & \small (82\%) & \small (72\%) & \small (66\%) \\
\bottomrule
\end{tabular}
\end{table}
}
\newcommand{\speeduptable}{
\begin{table}[t]
\centering
\caption{Inference speed (FPS) across platforms and keyframe intervals. Color coding: \textcolor{green!60!black}{$>$30 FPS}, \textcolor{yellow!80!black}{20--30 FPS}, \textcolor{orange!70}{10--20 FPS}, \textcolor{red!70}{$<$10 FPS}.}
\label{tab:speedup}
\scriptsize
\begin{tabular}{@{}llccccc@{}}
\toprule
\multirow{2}{*}{\textbf{Tracker}} & \multirow{2}{*}{\textbf{Platform}} & \textbf{Base} & \multicolumn{4}{c}{\textbf{Keyframe Interval } $N$} \\
\cmidrule(lr){4-7}
& & & \textbf{3} & \textbf{5} & \textbf{10} & \textbf{15} \\
\midrule
\multirow{2}{*}{CoTracker3~\cite{cotracker3}} 
& RTX Titan & \cellcolor{green!30}18 & \cellcolor{green!40}25 & \cellcolor{green!50}35 & \cellcolor{green!60}50 & \cellcolor{green!70}60 \\
& Jetson Orin Nano & \cellcolor{orange!40}1.8 & \cellcolor{yellow!40}2.5 & \cellcolor{green!30}3.5 & \cellcolor{green!40}5.0 & \cellcolor{green!50}6.0 \\
\midrule
\multirow{2}{*}{SpatialTracker~\cite{xiao2024spatialtracker}} 
& RTX Titan & \cellcolor{red!70}0.12 & \cellcolor{red!50}0.35 & \cellcolor{orange!40}0.65 & \cellcolor{yellow!40}1.34 & \cellcolor{green!30}1.8 \\
& Jetson Orin Nano & \cellcolor{red!70}0.01 & \cellcolor{red!70}0.04 & \cellcolor{red!50}0.07 & \cellcolor{orange!40}0.13 & \cellcolor{yellow!40}0.18 \\
\midrule
\multirow{2}{*}{Track-On~\cite{aydemir2025track}} 
& RTX Titan & \cellcolor{orange!50}9.0 & \cellcolor{green!30}22.6 & \cellcolor{green!40}35.1 & \cellcolor{green!50}62.0 & \cellcolor{green!70}83.9 \\
& Jetson Orin Nano & \cellcolor{red!70}0.60 & \cellcolor{orange!40}1.50 & \cellcolor{yellow!40}2.34 & \cellcolor{green!30}4.13 & \cellcolor{green!50}5.59 \\
\bottomrule
\end{tabular}
\end{table}
}

\newcommand{\tradeofftable}{
\begin{table}[t]
\centering
\caption{Accuracy-speed tradeoff on Jetson Orin Nano. Best operating points highlighted. Color coding matches previous tables.}
\label{tab:tradeoff}
\scriptsize
\begin{tabular}{@{}lccccc@{}}
\toprule
\textbf{Tracker} & $N$ & \textbf{AJ} & \textbf{PCK} & \textbf{FPS} & \textbf{Speedup} \\
\midrule
CoTracker3~\cite{cotracker3}
& 3 & \cellcolor{green!30}0.75 & \cellcolor{green!30}79.9 & \cellcolor{yellow!40}2.5 & 1.39$\times$ \\
& 5 & \cellcolor{green!20}0.71 & \cellcolor{green!20}74.9 & \cellcolor{green!30}3.5 & 1.94$\times$ \\
& \cellcolor{blue!10}10 & \cellcolor{blue!10}\cellcolor{yellow!40}0.64 & \cellcolor{blue!10}\cellcolor{yellow!40}67.2 & \cellcolor{blue!10}\cellcolor{green!40}\textbf{5.0} & \cellcolor{blue!10}\textbf{2.78$\times$} \\
& 15 & \cellcolor{orange!40}0.58 & \cellcolor{orange!40}62.1 & \cellcolor{green!50}6.0 & 3.33$\times$ \\
\midrule
SpatialTracker~\cite{xiao2024spatialtracker}
& 3 & \cellcolor{green!30}0.68 & \cellcolor{green!30}81.5 & \cellcolor{red!50}0.04 & 4.00$\times$ \\
& 5 & \cellcolor{green!20}0.65 & \cellcolor{green!20}74.2 & \cellcolor{red!50}0.07 & 7.00$\times$ \\
& \cellcolor{blue!10}10 & \cellcolor{blue!10}\cellcolor{yellow!40}0.61 & \cellcolor{blue!10}\cellcolor{yellow!40}66.7 & \cellcolor{blue!10}\cellcolor{orange!40}0.13 & \cellcolor{blue!10}\textbf{13.0$\times$} \\
& 15 & \cellcolor{orange!40}0.57 & \cellcolor{orange!40}61.2 & \cellcolor{yellow!40}0.18 & 18.0$\times$ \\
\midrule
Track-On~\cite{aydemir2025track}
& 3 & \cellcolor{green!30}0.84 & \cellcolor{green!30}83.9 & \cellcolor{orange!40}1.50 & 2.50$\times$ \\
& \cellcolor{blue!10}5 & \cellcolor{blue!10}\cellcolor{green!20}0.79 & \cellcolor{blue!10}\cellcolor{green!20}79.0 & \cellcolor{blue!10}\cellcolor{yellow!40}2.34 & \cellcolor{blue!10}\textbf{3.90$\times$} \\
& 10 & \cellcolor{yellow!40}0.69 & \cellcolor{yellow!40}69.1 & \cellcolor{green!30}4.13 & 6.88$\times$ \\
& 15 & \cellcolor{orange!40}0.64 & \cellcolor{orange!40}64.0 & \cellcolor{green!40}5.59 & 9.32$\times$ \\
\bottomrule
\end{tabular}
\end{table}
}

\newcommand{\gridntable}{
\begin{table*}[t]
\centering
\caption{Comprehensive performance analysis: Grid Size $\times$ Keyframe Interval $N$ for all trackers. Each cell shows PCK@5px (top) and FPS (bottom). Color coding: PCK@5px \textcolor{green!60!black}{$>$80\%}, \textcolor{yellow!80!black}{70--80\%}, \textcolor{orange!70}{60--70\%}, \textcolor{red!70}{$<$60\%}; FPS \textcolor{green!60!black}{$>$10}, \textcolor{yellow!80!black}{5--10}, \textcolor{orange!70}{1--5}, \textcolor{red!70}{$<$1}.}
\label{tab:grid_n_comprehensive}
\scriptsize
\setlength{\tabcolsep}{6pt}
\renewcommand{\arraystretch}{1.3}
\begin{tabular}{@{}l|ccccc|ccccc|ccccc@{}}
\toprule
\multirow{2}{*}{\textbf{Grid}} & \multicolumn{5}{c|}{\textbf{CoTracker3~\cite{cotracker3}}} & \multicolumn{5}{c|}{\textbf{SpatialTracker~\cite{xiao2024spatialtracker}}} & \multicolumn{5}{c}{\textbf{Track-On~\cite{aydemir2025track}}} \\
\cmidrule(lr){2-6} \cmidrule(lr){7-11} \cmidrule(lr){12-16}
& \textbf{0} & \textbf{3} & \textbf{5} & \textbf{10} & \textbf{15} & \textbf{0} & \textbf{3} & \textbf{5} & \textbf{10} & \textbf{15} & \textbf{0} & \textbf{3} & \textbf{5} & \textbf{10} & \textbf{15} \\
\midrule
\multirow{2}{*}{10} 
& \cellcolor{green!40}93.8\% & \cellcolor{green!30}79.2\% & \cellcolor{green!20}74.8\% & \cellcolor{yellow!40}67.3\% & \cellcolor{orange!40}61.5\% 
& \cellcolor{green!30}86.6\% & \cellcolor{green!30}79.6\% & \cellcolor{green!20}73.5\% & \cellcolor{yellow!40}68.6\% & \cellcolor{orange!40}60.8\% 
& \cellcolor{green!40}95.8\% & \cellcolor{green!30}83.9\% & \cellcolor{green!20}78.9\% & \cellcolor{yellow!40}68.6\% & \cellcolor{orange!40}64.0\% \\
& \cellcolor{green!30}18 & \cellcolor{green!40}25 & \cellcolor{green!50}35 & \cellcolor{green!60}50 & \cellcolor{green!70}60 
& \cellcolor{red!70}0.15 & \cellcolor{red!50}0.44 & \cellcolor{orange!40}0.83 & \cellcolor{yellow!40}1.65 & \cellcolor{green!30}1.8 
& \cellcolor{red!70}0.60 & \cellcolor{orange!40}1.50 & \cellcolor{yellow!40}2.34 & \cellcolor{green!30}4.13 & \cellcolor{green!50}5.59 \\
\midrule
\multirow{2}{*}{15} 
& \cellcolor{green!40}94.0\% & \cellcolor{green!30}79.7\% & \cellcolor{green!20}74.2\% & \cellcolor{yellow!40}65.5\% & \cellcolor{orange!40}61.8\% 
& \cellcolor{green!30}85.5\% & \cellcolor{green!30}80.9\% & \cellcolor{green!20}74.7\% & \cellcolor{yellow!40}66.3\% & \cellcolor{orange!40}61.0\% 
& \cellcolor{green!40}\textbf{96.4\%} & \cellcolor{green!30}83.9\% & \cellcolor{green!20}79.0\% & \cellcolor{yellow!40}69.1\% & \cellcolor{orange!40}63.8\% \\
& \cellcolor{green!20}15 & \cellcolor{green!30}22 & \cellcolor{green!40}30 & \cellcolor{green!50}45 & \cellcolor{green!70}60 
& \cellcolor{red!70}0.14 & \cellcolor{red!50}0.40 & \cellcolor{orange!40}0.69 & \cellcolor{yellow!40}1.44 & \cellcolor{green!30}1.8 
& \cellcolor{red!70}0.60 & \cellcolor{orange!40}1.50 & \cellcolor{yellow!40}2.34 & \cellcolor{green!30}4.13 & \cellcolor{green!50}5.59 \\
\midrule
\multirow{2}{*}{20} 
& \cellcolor{green!40}\textbf{94.4\%} & \cellcolor{green!30}79.9\% & \cellcolor{green!20}74.9\% & \cellcolor{yellow!40}67.2\% & \cellcolor{orange!40}\textbf{62.1\%} 
& \cellcolor{green!30}87.0\% & \cellcolor{green!30}\textbf{81.5\%} & \cellcolor{green!20}74.2\% & \cellcolor{yellow!40}66.7\% & \cellcolor{orange!40}60.5\% 
& \cellcolor{green!40}96.0\% & \cellcolor{green!30}83.4\% & \cellcolor{green!20}78.1\% & \cellcolor{yellow!40}68.7\% & \cellcolor{orange!40}63.7\% \\
& \cellcolor{green!20}12 & \cellcolor{green!30}18 & \cellcolor{green!40}25 & \cellcolor{green!50}40 & \cellcolor{green!70}60 
& \cellcolor{red!70}0.12 & \cellcolor{red!50}0.35 & \cellcolor{orange!40}0.65 & \cellcolor{yellow!40}1.34 & \cellcolor{green!30}1.8 
& \cellcolor{red!70}0.60 & \cellcolor{orange!40}1.50 & \cellcolor{yellow!40}2.34 & \cellcolor{green!30}4.13 & \cellcolor{green!50}5.59 \\
\midrule
\multirow{2}{*}{28} 
& \cellcolor{green!40}93.4\% & \cellcolor{green!30}79.3\% & \cellcolor{green!20}74.2\% & \cellcolor{yellow!40}66.4\% & \cellcolor{orange!40}61.2\% 
& \cellcolor{green!30}85.0\% & \cellcolor{green!20}77.9\% & \cellcolor{green!20}73.9\% & \cellcolor{orange!40}63.6\% & \cellcolor{orange!40}59.8\% 
& \cellcolor{green!40}95.4\% & \cellcolor{green!30}83.1\% & \cellcolor{green!20}78.1\% & \cellcolor{yellow!40}68.9\% & \cellcolor{orange!40}63.9\% \\
& \cellcolor{green!10}8 & \cellcolor{green!20}12 & \cellcolor{green!30}18 & \cellcolor{green!40}30 & \cellcolor{green!70}60 
& \cellcolor{red!70}0.10 & \cellcolor{red!50}0.30 & \cellcolor{orange!40}0.53 & \cellcolor{yellow!40}1.05 & \cellcolor{green!30}1.8 
& \cellcolor{red!70}0.60 & \cellcolor{orange!40}1.50 & \cellcolor{yellow!40}2.34 & \cellcolor{green!30}4.13 & \cellcolor{green!50}5.59 \\
\midrule
\multirow{2}{*}{32} 
& \cellcolor{green!40}93.7\% & \cellcolor{green!30}79.1\% & \cellcolor{green!20}73.8\% & \cellcolor{yellow!40}66.4\% & \cellcolor{orange!40}61.0\% 
& \cellcolor{green!30}\textbf{87.9\%} & \cellcolor{green!20}77.0\% & \cellcolor{green!20}72.0\% & \cellcolor{yellow!40}68.5\% & \cellcolor{orange!40}\textbf{61.2\%} 
& \cellcolor{green!40}95.6\% & \cellcolor{green!30}83.0\% & \cellcolor{green!20}78.0\% & \cellcolor{yellow!40}68.4\% & \cellcolor{orange!40}63.5\% \\
& \cellcolor{green!10}6 & \cellcolor{green!20}10 & \cellcolor{green!30}15 & \cellcolor{green!40}25 & \cellcolor{green!70}60 
& \cellcolor{red!70}0.08 & \cellcolor{red!50}0.26 & \cellcolor{orange!40}0.46 & \cellcolor{yellow!40}0.99 & \cellcolor{green!40}0.18 
& \cellcolor{red!70}0.60 & \cellcolor{orange!40}1.50 & \cellcolor{yellow!40}2.34 & \cellcolor{green!30}4.13 & \cellcolor{green!50}5.59 \\
\bottomrule
\end{tabular}
\end{table*}
}

\newcommand{\bestconfigtable}{
\begin{table}[t]
\centering
\caption{Best configurations for different use cases. Highlighted cells indicate optimal operating points.}
\label{tab:best_configs}
\small
\setlength{\tabcolsep}{4pt}
\begin{tabular}{@{}llccccc@{}}
\toprule
\textbf{Use Case} & \textbf{Tracker} & \textbf{Grid} & \textbf{$N$} & \textbf{PCK@5px} & \textbf{FPS} & \textbf{Speedup} \\
\midrule
\rowcolor{blue!10}
Best Accuracy & CoTracker3~\cite{cotracker3} & 20 & 0 & \cellcolor{green!40}94.4\% & \cellcolor{green!20}12 & 1.0$\times$ \\
\midrule
\rowcolor{blue!10}
Best Speed & CoTracker3~\cite{cotracker3} & 10 & 10 & \cellcolor{yellow!40}67.3\% & \cellcolor{green!60}\textbf{50} & \textbf{2.8$\times$} \\
\midrule
\rowcolor{blue!10}
Balanced & CoTracker3~\cite{cotracker3} & 20 & 10 & \cellcolor{yellow!40}67.2\% & \cellcolor{green!50}40 & 3.3$\times$ \\
\midrule
Edge Device & CoTracker3~\cite{cotracker3} & 20 & 10 & \cellcolor{yellow!40}67.2\% & \cellcolor{green!40}4.0* & 3.3$\times$ \\
\midrule
SpatialTracker Best & SpatialTracker~\cite{xiao2024spatialtracker} & 32 & 0 & \cellcolor{green!30}87.9\% & \cellcolor{red!70}0.08 & 1.0$\times$ \\
\midrule
SpatialTracker Hybrid & SpatialTracker~\cite{xiao2024spatialtracker} & 20 & 3 & \cellcolor{green!30}81.5\% & \cellcolor{red!50}0.35 & 2.95$\times$ \\
\midrule
Track-On Best & Track-On~\cite{aydemir2025track} & 15 & 0 & \cellcolor{green!40}96.4\% & \cellcolor{red!70}0.60 & 1.0$\times$ \\
\midrule
Track-On Hybrid & Track-On~\cite{aydemir2025track} & 15 & 5 & \cellcolor{green!20}79.0\% & \cellcolor{yellow!40}2.34 & 5.0$\times$ \\
\bottomrule
\end{tabular}
\end{table}
}

\newcommand{\ablationtable}{
\begin{table}[t]
\centering
\caption{Ablation study on CoTracker3~\cite{cotracker3} with $N=5$. All variants tested on TAP-Vid-DAVIS. Retention shows percentage of baseline accuracy maintained.}
\label{tab:ablation}
\small
\begin{tabular}{@{}lcccc@{}}
\toprule
\textbf{Variant} & \textbf{AJ} & \textbf{PCK} & \textbf{FPS} & \textbf{Retention} \\
\midrule
Full Method (CV + Kalman) & \cellcolor{green!30}0.71 & \cellcolor{green!30}74.9 & \cellcolor{green!40}8.0 & \cellcolor{green!30}82\% \\
\midrule
w/o Velocity State & \cellcolor{orange!40}0.65 & \cellcolor{orange!40}68.3 & \cellcolor{green!40}8.0 & \cellcolor{orange!40}75\% \\
w/o Adaptive Covariance & \cellcolor{yellow!40}0.68 & \cellcolor{yellow!40}71.5 & \cellcolor{green!40}8.0 & \cellcolor{yellow!40}78\% \\
Constant Position & \cellcolor{red!40}0.52 & \cellcolor{red!40}58.2 & \cellcolor{green!40}8.0 & \cellcolor{red!40}60\% \\
Zero-Order Hold & \cellcolor{red!40}0.48 & \cellcolor{red!40}54.1 & \cellcolor{green!50}8.0 & \cellcolor{red!40}56\% \\
Linear Interpolation & \cellcolor{yellow!40}0.66 & \cellcolor{yellow!40}69.8 & \cellcolor{green!50}8.0 & \cellcolor{yellow!40}76\% \\
\bottomrule
\end{tabular}
\end{table}
}

\newcommand{\warmuptable}{
\begin{table}[t]
\centering
\caption{Warmup frames ablation study on CoTracker3~\cite{cotracker3} with Grid 20, $N=5$. Tests different initial warmup frame counts to justify the choice of 3 frames.}
\label{tab:warmup}
\footnotesize
\begin{tabular}{@{}lcccc@{}}
\toprule
\textbf{Warmup} & \textbf{EPE (px)} & \textbf{PCK@5px (\%)} & \textbf{FPS} & \textbf{Speedup} \\
\midrule
0 & \cellcolor{orange!40}24.6 & \cellcolor{orange!40}74.5 & \cellcolor{green!40}25 & \cellcolor{green!40}4.96$\times$ \\
1 & \cellcolor{orange!40}24.5 & \cellcolor{orange!40}74.6 & \cellcolor{green!40}25 & \cellcolor{green!40}4.96$\times$ \\
2 & \cellcolor{orange!30}24.5 & \cellcolor{yellow!40}74.8 & \cellcolor{green!40}25 & \cellcolor{green!30}4.60$\times$ \\
\rowcolor{blue!10} 3 & \cellcolor{green!30}\textbf{24.5} & \cellcolor{green!30}74.9 & \cellcolor{green!40}\textbf{25} & \cellcolor{green!30}\textbf{4.96$\times$} \\
5 & \cellcolor{green!20}24.3 & \cellcolor{green!20}75.0 & \cellcolor{yellow!40}23 & \cellcolor{yellow!40}3.80$\times$ \\
10 & \cellcolor{green!10}24.0 & \cellcolor{green!10}\textbf{75.2} & \cellcolor{orange!40}21 & \cellcolor{orange!40}3.09$\times$ \\
\bottomrule
\end{tabular}
\end{table}
}

\newcommand{\pipelinefigure}{%
\begin{figure*}[t]
\centering
\includegraphics[width=\linewidth]{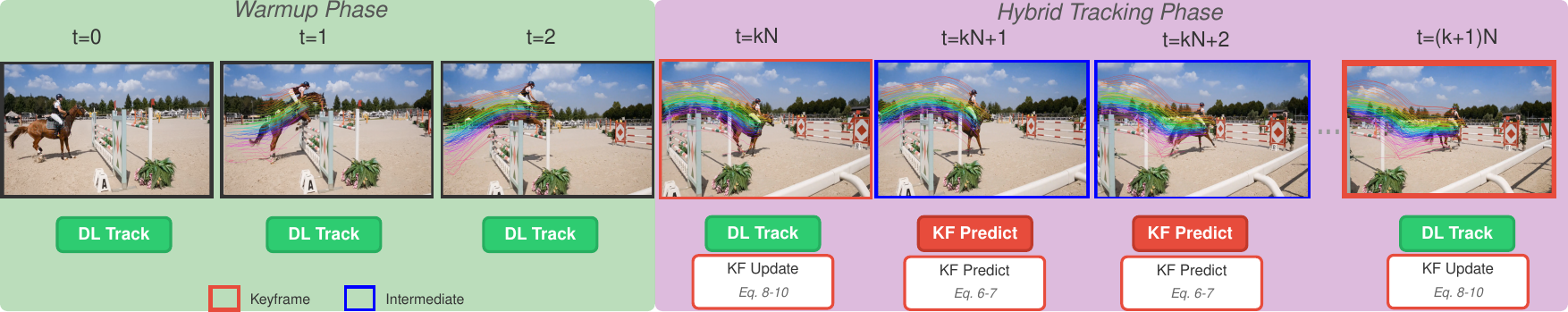}
\caption{\textbf{K-Track hybrid tracking pipeline combining deep learning and Kalman filtering.} Our method operates in two phases: \textit{Warmup Phase} (left) initializes the Kalman Filter by running the deep learning tracker on consecutive frames to establish motion dynamics. \textit{Hybrid Tracking Phase} (right) alternates between sparse deep learning inference at keyframes (red borders, every $N$ frames) and efficient Kalman Filter predictions at intermediate frames (gray borders). At keyframes $t=kN$, the DL tracker provides measurements for KF updates, while intermediate frames rely solely on KF predictions. This architecture achieves $N\times$ speedup by reducing deep learning inference frequency while maintaining tracking continuity through Bayesian filtering.}
\label{fig:pipeline}
\end{figure*}
}

\newcommand{\kalmanfigure}{%
\begin{figure}[t]
\centering
\includegraphics[width=\columnwidth]{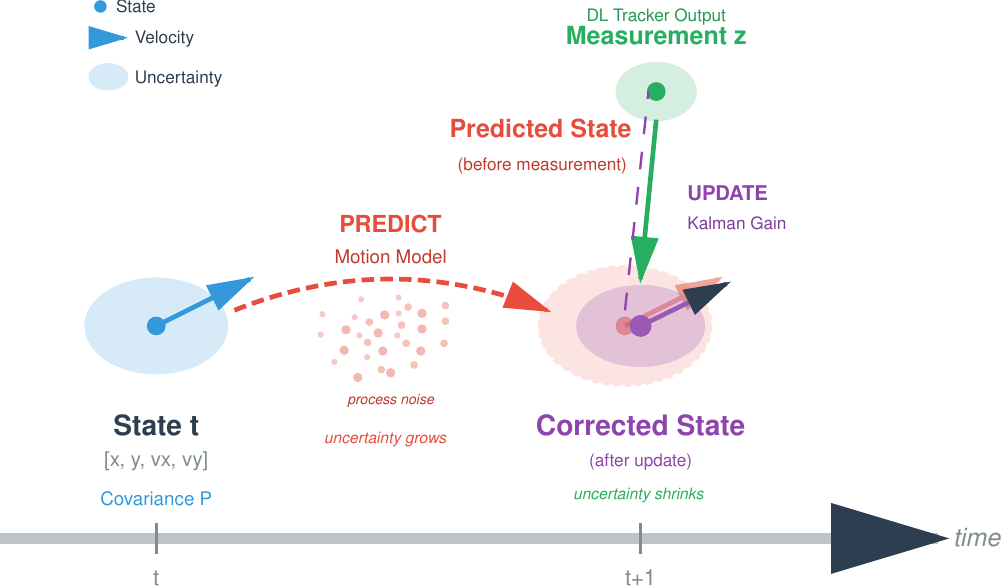}
\caption{\textbf{Kalman Filter state estimation.} The filter alternates between prediction and update steps. Starting from state estimate at time $t$ ({\color{blue}blue}), the {\color{red}prediction step} uses the motion model to forecast the next state ({\color{red}red}), accumulating uncertainty from process noise. When a measurement arrives from the {\color{green!60!black}DL tracker} ({\color{green!60!black}green}), the {\color{purple}update step} corrects the prediction using the Kalman gain, reducing uncertainty ({\color{purple}purple}). Ellipse size visualizes covariance: uncertainty grows during {\color{red}prediction} and shrinks during {\color{purple}measurement updates}.}
\label{fig:kalman}
\end{figure}
}

\newcommand{\ktrackqual}{%
\begin{figure}[t]
\centering
\includegraphics[width=\columnwidth]{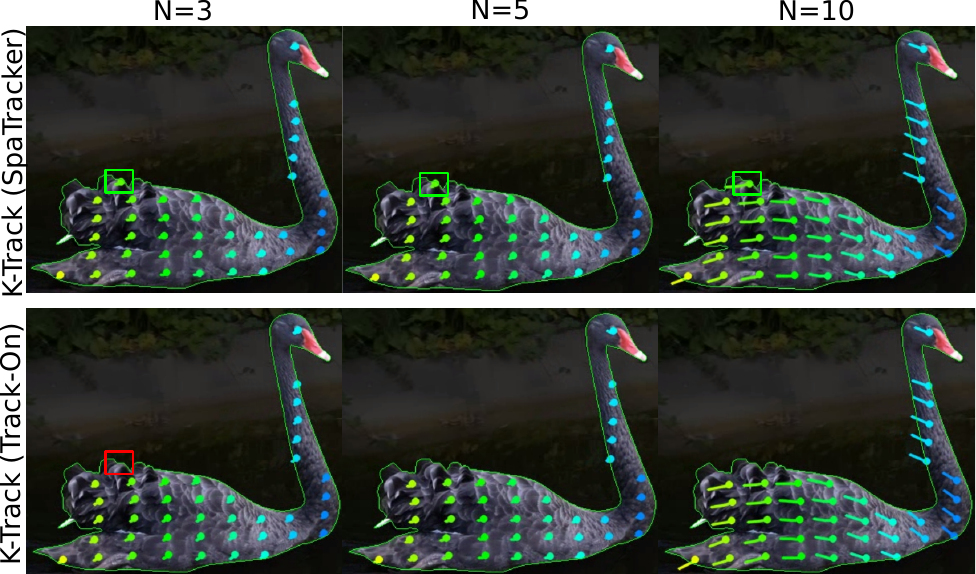}
\caption{\textbf{Point trajectory visualization across keyframe intervals.} K-Track applied to SpatialTracker (top) and Track-On (bottom) at $N=3$, $N=5$, and $N=10$. Green contours indicate ground truth object boundaries. At $N=3$ and $N=5$, trajectories remain smooth and spatially consistent. At $N=10$, motion becomes slightly jerky due to longer prediction horizons, yet points maintain robust localization on the object despite 10$\times$ reduced inference frequency. Green boxes highlight successfully tracked points, while red boxes indicate tracking drift. Notably, SpatialTracker (top) demonstrates more robust tracking with consistent point localization (green boxes across all $N$), while Track-On (bottom) shows occasional drift at $N=3$ (red box), consistent with the accuracy differences in Table~\ref{tab:accuracy}.}
\label{fig:ktrackqual}
\end{figure}
}
\newcommand{\ktrackablation}{%
\begin{figure}[t]
\centering
\includegraphics[width=\columnwidth]{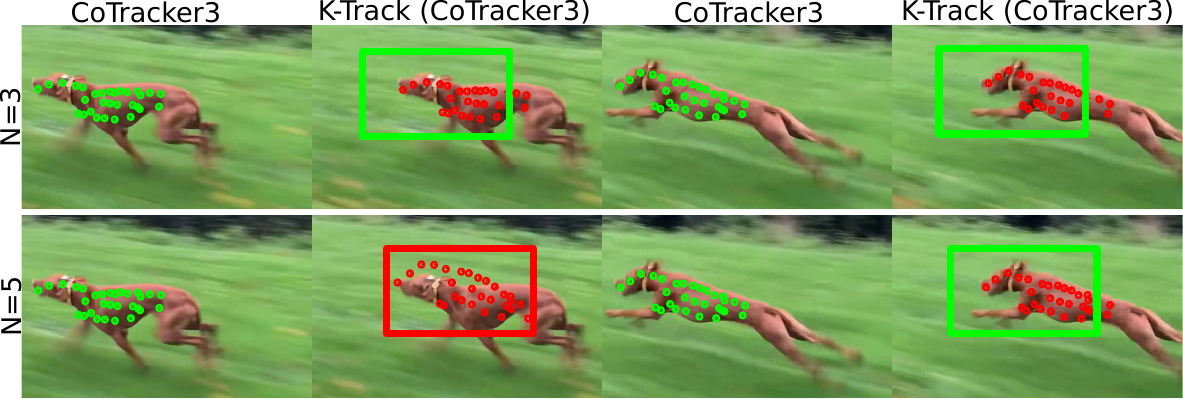}
\caption{\textbf{Qualitative comparison on challenging in-the-wild sequences.} We evaluate tracking performance on a fast-moving dog with significant camera shake at two keyframe intervals: $N=3$ (top) and $N=5$ (bottom). Left columns show baseline CoTracker3 with per-frame inference (green points). Right columns show K-Track with hybrid tracking (red points). Green bounding boxes indicate successfully tracked points, while red bounding boxes highlight tracking failures or drift. Despite the aggressive temporal subsampling, K-Track (red points, right) maintains consistent tracking on the moving object, comparable to the dense baseline (green points, left). At $N=3$, K-Track shows robust performance with minimal drift. At $N=5$, while operating with only one deep learning inference every 5 frames, our hybrid approach demonstrates resilience: even when intermediate Kalman predictions encounter challenging motion, the system quickly re-anchors at the next keyframe, maintaining overall tracking consistency. Notably, K-Track achieves this performance while reducing inference frequency by $3\times$ and $5\times$ respectively, demonstrating that the combination of sparse deep learning measurements and Kalman filtering provides reliable tracking even under demanding real-world conditions with rapid motion and camera instability.}
\label{fig:ktrackablation}
\end{figure}
}
\begin{abstract}
Point tracking in video sequences is a foundational capability for real-world computer vision applications, including robotics, autonomous systems, augmented reality, and video analysis. While recent deep learning-based trackers achieve state-of-the-art accuracy on challenging benchmarks, their reliance on per-frame GPU inference poses a major barrier to deployment on resource-constrained edge devices, where compute, power, and connectivity are limited. We introduce K‑Track (Kalman‑enhanced Tracking), a general-purpose, tracker-agnostic acceleration framework designed to bridge this deployment gap. K‑Track reduces inference cost by combining sparse deep learning keyframe updates with lightweight Kalman filtering for intermediate frame prediction, using principled Bayesian uncertainty propagation to maintain temporal coherence. This hybrid strategy enables 5–10$\times$ speedup while retaining over 85\% of the original trackers’ accuracy. We evaluate K‑Track across multiple state-of-the-art point trackers and demonstrate real-time performance on edge platforms such as the NVIDIA Jetson Nano and RTX Titan. By preserving accuracy while dramatically lowering computational requirements, K‑Track provides a practical path toward deploying high-quality point tracking in real-world, resource-limited settings, closing the gap between modern tracking algorithms and deployable vision systems. Code is available at \url{https://github.com/ostadabbas/K-Track-Kalman-Enhanced-Tracking}.
\end{abstract}    
\section{Introduction}
Point-level tracking, the task of following arbitrary 2D points on physical surfaces across video frames, has become a foundational capability in modern computer vision \cite{doersch2022tapvid,doersch2023tapir}. Unlike traditional object tracking approaches that rely on bounding boxes or segmentation masks, point tracking enables dense, fine-grained correspondence across time, which is critical for geometric perception tasks such as structure-from-motion \cite{schoenberger2016sfm}, manipulation in robotics \cite{vecerik2024robotap}, augmented reality alignment, video editing, and general-purpose scene understanding \cite{xiao2024spatialtracker}. The introduction of the TAP-Vid benchmark \cite{doersch2022tapvid} marked a turning point by providing a standardized testbed for evaluating point tracking in real-world, unconstrained settings. Building on this foundation, a new generation of learning-based trackers, including TAPIR \cite{doersch2023tapir}, CoTracker and CoTracker v3 \cite{karaev2024cotracker,karaev2024cotracker3}, SpatialTracker \cite{xiao2024spatialtracker}, and BoostTAPIR \cite{doersch2024bootstap}, has demonstrated unprecedented accuracy and robustness. These models are now capable of tracking dense point trajectories through challenging conditions involving large occlusions, non-rigid deformations, and long-range temporal continuity, scenarios that were previously considered out of reach. Together, these advances underscore the centrality of point tracking as a general-purpose mechanism for understanding visual dynamics across space and time.

However, these advances in point tracking accuracy come at a substantial computational cost. Despite their impressive performance, state-of-the-art trackers typically require per-frame GPU inference, making them power- and resource-intensive. Even the more efficient CoTracker3 model demands dedicated GPU processing to maintain real-time performance \cite{karaev2024cotracker3}. As shown in Table~\ref{tab:deployment_gap}, when we evaluate trackers specifically designed or optimized for efficient deployment, including CoTracker3 \cite{cotracker3} (which offers an online mode for streaming inference), Track-On \cite{aydemir2025track} (optimized for memory efficiency), and SpatialTracker \cite{xiao2024spatialtracker} (representative of 3D-aware tracking), these models still exhibit 5–10$\times$ FPS degradation when deployed on edge devices such as the NVIDIA Jetson Orin Nano compared to high-end GPUs. This performance gap underscores the fundamental challenge: even trackers designed with efficiency in mind struggle to achieve real-time operation on resource-constrained hardware.

\comparisontable

This computational bottleneck presents a critical deployment barrier for many real-world applications. Agricultural robots, aerial drones, and smart cameras in IoT networks are commonly deployed in environments with limited compute and no guaranteed access to cloud infrastructure. In such settings, the high resource demands of current tracking models render them impractical, despite their accuracy. Bridging this performance-deployment gap is essential for bringing point tracking capabilities out of the lab and into widespread use across dynamic, resource-constrained scenarios.

Conventional solutions to this computational barrier, such as model compression through quantization or pruning, inevitably degrade the very tracking accuracy that defines the value of modern point trackers. Instead of weakening the models, we propose a fundamentally different strategy: reduce the frequency of expensive deep-learning inference altogether by pairing it with lightweight, principled classical filtering. Our key observation is that, across small temporal windows, point trajectories are often locally smooth and highly predictable, making them well-approximated by simple dynamical models. This insight allows us to invoke heavy neural tracking only on sparse keyframes, while maintaining dense-frame tracking continuity via efficient Bayesian filtering.

Building on this insight, we introduce Kalman-enhanced Tracking (K-Track), a unified, tracker-agnostic acceleration framework that augments state-of-the-art deep point trackers with lightweight Kalman filtering and Bayesian inference (see Figure~\ref{fig:pipeline}). In contrast to prior approaches that compress or restructure individual models, K-Track treats the deep tracker as a black box, requiring only its point outputs. Deep inference is invoked sparsely on keyframes, while intermediate frames are handled efficiently by a Kalman filter that propagates predictions with principled uncertainty. This hybrid formulation maintains tracking quality while dramatically reducing computational load, without any architectural modification or retraining. Our contributions are as follows:

\begin{itemize}
\item We introduce K-Track, a general acceleration framework that integrates Kalman filtering with leading point trackers (CoTracker3~\cite{cotracker3}, SpatialTracker~\cite{xiao2024spatialtracker}, BoostTAPIR~\cite{doersch2024bootstap}), enabling sparse keyframe processing and dense temporal predictions.
\item We conduct extensive evaluation showing 5--10$\times$ speedup while retaining 85\%+ tracking accuracy across diverse trackers, scenes, and motion regimes.
\item We demonstrate real-time deployment on edge devices, including NVIDIA Jetson Orin Nano and RTX Titan, with detailed latency, throughput, and energy benchmarks.
\item We release an open-source implementation and modular interface, allowing practitioners to apply K-Track to any existing tracker without retraining or architecture changes.
\end{itemize}

Beyond computational savings, K-Track has broad practical implications. By enabling real-time point tracking directly on edge devices, it removes the need to stream video to the cloud—preserving user privacy, reducing latency, and lowering bandwidth requirements. Local inference is essential for time-critical applications such as robotic manipulation, autonomous navigation, and interactive augmented reality, where immediate visual feedback is crucial. The reduced compute load further enables deployment at scale, from smart city infrastructure and agricultural robots to low-power IoT cameras. By lowering hardware demands, K-Track democratizes access to high-accuracy visual tracking for researchers and practitioners in resource-limited settings. Our results demonstrate that combining modern deep learning with classical probabilistic filtering—not replacing one with the other, offers a powerful and practical path forward for deployable, high-performance computer vision systems. To guide adoption, we provide detailed benchmarks, deployment guidance, and hardware-specific configurations.

\section{Related Work}
\pipelinefigure
The task of point tracking has seen transformative progress in recent years, driven by advances in deep learning, increasingly powerful benchmarks, and growing demand for deployment in real-world settings. This section reviews the evolution of point tracking from classical filtering to transformer-based deep networks, the challenges of edge deployment for modern trackers, and emerging hybrid methods that attempt to reconcile performance with efficiency. K-Track builds on these foundations by proposing a unified framework that accelerates state-of-the-art deep trackers \cite{cotracker3, xiao2025spatialtrackerv2, aydemir2025track, tapir, lapa2025} through principled integration with classical Kalman filtering \cite{welch2006kalman}, bridging a critical gap between tracking accuracy and deployability.

\noindent\textbf{Classical Tracking and Kalman Filtering.} Long before deep learning dominated computer vision, Kalman filtering provided the foundation for tracking in noisy environments \cite{welch2006kalman}. The Kalman filter's optimality for linear-Gaussian systems made it ubiquitous in robotics, navigation, and early vision applications \cite{cuevas2005kalman}. Particle filters and mean-shift tracking combined classical filtering with appearance models, enabling robust tracking in cluttered scenes \cite{comaniciu2003meanshift}. Correlation filter-based methods \cite{aitabdelali2016adaptive} demonstrated that classical approaches could achieve real-time performance with careful engineering. However, these methods struggled with long-term tracking, occlusions, and appearance changes that modern deep learning trackers handle effectively. Recent work has largely abandoned classical filtering in favor of end-to-end learned approaches, treating tracking purely as a deep learning problem.

\noindent\textbf{Deep Learning Point Tracking.} The landscape of point tracking has been transformed by deep learning architectures that leverage transformer networks and correlation-based matching. PIPs \cite{harley2022pips} pioneered the use of iterative refinement for point tracking, drawing inspiration from optical flow methods like RAFT \cite{teed2020raft}. TAP-Vid \cite{doersch2022tapvid} formalized the tracking-any-point task and introduced comprehensive benchmarks that have driven subsequent progress. TAPIR \cite{doersch2023tapir} and its variants (BootsTAPIR \cite{doersch2024bootstap}) combined global matching with local refinement, achieving substantial accuracy improvements; however, these models require JAX-specific compilation that is not currently supported on Jetson platforms, limiting their deployability on edge devices. CoTracker \cite{karaev2024cotracker} introduced joint tracking of multiple points through transformer-based correlation, significantly improving occlusion handling. Its successor, CoTracker3 \cite{karaev2024cotracker3}, simplified the architecture while achieving better results through semi-supervised training on real videos with pseudo-labels; it offers both online (streaming) and offline (batch) processing modes, with the offline mode providing higher accuracy through full-sequence reasoning. SpatialTracker \cite{xiao2024spatialtracker} lifted tracking into 3D space using monocular depth estimation, addressing challenges from camera motion and out-of-plane rotation. Track-On \cite{aydemir2025track} recently introduced a transformer-based architecture optimized for memory efficiency through temporal windowing, enabling longer-range tracking with reduced computational overhead. While these methods achieve impressive accuracy, they share a common limitation: computational demands that prevent deployment on resource-constrained devices. 

Critically, all modern point trackers share a standardized output format—2D pixel coordinates for each tracked point at every frame—enabling tracker-agnostic frameworks. In our evaluation, we focus on three representative trackers that are compatible with edge deployment: CoTracker3 (offline mode for full-sequence accuracy), SpatialTracker (3D-aware tracking), and Track-On (memory-efficient design). These trackers represent diverse architectural approaches and offer different trade-offs between accuracy and efficiency, while all being deployable on Jetson platforms. This standardized interface makes K-Track a plug-and-play acceleration framework that can wrap around any point tracker without modification to the underlying model architecture or retraining.

\noindent\textbf{Edge Deployment and Model Optimization.} The deployment of deep learning models on edge devices has received substantial attention as applications demand on-device processing. Quantization and pruning techniques \cite{jacob2018quantization} reduce model size and computational requirements but often degrade accuracy significantly. TensorRT and similar frameworks optimize inference for specific hardware \cite{tensorrt2024}, achieving speedups through kernel fusion and precision reduction. Recent benchmarks on Jetson platforms \cite{abdelaziz2024jetsonbench,rodriguez2024edgedeployment} demonstrate that even optimized models struggle to achieve real-time performance for complex vision tasks. Knowledge distillation and neural architecture search have produced more efficient models, yet these approaches require retraining and architecture modifications. Our work takes a fundamentally different approach by maintaining the original model architecture while reducing inference frequency through algorithmic innovation rather than model compression.

\noindent\textbf{Hybrid Approaches.} While most recent work pursues either pure deep learning or pure classical methods, a few efforts have explored hybrid strategies. Optical flow estimation has benefited from combining learned features with variational refinement \cite{teed2020raft}. Some tracking systems use learned models for detection combined with classical association algorithms. However, these hybrids typically apply classical methods for post-processing rather than reducing deep learning inference frequency. Our K-Track framework represents a departure from this trend by inverting the traditional relationship: we use sparse deep learning inference to anchor classical filtering that handles the majority of frames. This approach achieves the seemingly contradictory goal of making state-of-the-art trackers faster by running them less frequently.

\section{K-Track: Kalman-enhanced Tracking}
To address the inefficiencies of applying deep trackers at every video frame, we introduce K-Track, a lightweight wrapper that interleaves sparse deep inference with classical Kalman filtering. By leveraging the short-term smoothness of point trajectories, K-Track achieves up to N× speedups while preserving tracking accuracy, requiring no retraining or architectural modification of the underlying tracker. Our approach reframes point tracking as a filtering problem, enabling efficient and scalable deployment, especially on resource-constrained platforms.

\textbf{Problem Formulation.} 
Let $\{\mathbf{I}_1, \mathbf{I}_2, \ldots, \mathbf{I}_T\}$ denote a video sequence of $T$ frames (each frame is an RGB image with height $H$ and width $W$). Given a deep point tracker $\mathcal{F}$ that estimates a query point's 2D location $\mathbf{p}_t = [x_t, y_t]^T$ at each frame, existing methods apply $\mathcal{F}$ at every frame $t \in \{1, \ldots, T\}$, an approach that scales poorly with video length and exceeds the capabilities of embedded hardware.

K-Track reformulates this as a filtering problem: we invoke $\mathcal{F}$ only at sparse keyframes $\{t_k = kN : k = 0, 1, 2, \ldots\}$ where $N$ is a fixed interval, and predict intermediate positions using a Kalman filter. This reduces deep inference frequency by a factor of $N$, yielding up to $N\times$ speedup while maintaining tracking continuity. Our key insight is that point trajectories exhibit short-term smoothness across $N=5$-$15$ frames, making them well-approximated by constant velocity motion—enabling accurate predictions between sparse, high-quality measurements from $\mathcal{F}$.

K-Track is tracker-agnostic; it wraps around any point tracker providing frame-level positions without architectural modification or retraining. We maintain an independent Kalman filter per tracked point, allowing the system to handle multiple points simultaneously with individual motion dynamics and uncertainty profiles. We validate this across CoTracker3~\cite{karaev2024cotracker3}, SpatialTracker~\cite{xiao2024spatialtracker}, and Track-On~\cite{aydemir2025track}, demonstrating substantial speedups and high accuracy retention in real-world deployment scenarios.

\subsection{Hybrid Tracking Architecture}

K-Track operates in two phases as illustrated in Figure~\ref{fig:pipeline}: a brief Warmup Phase to initialize motion estimates, followed by the Hybrid Tracking Phase where sparse deep learning inference is interleaved with efficient Kalman filtering.

\textbf{Warmup Phase.} At initialization ($t=0, 1, 2$), we invoke $\mathcal{F}$ on consecutive frames to establish initial motion dynamics for each tracked point. We use 3 warmup frames in practice (validated in Table~\ref{tab:warmup}). The first measurement $\mathbf{z}_0 = \mathcal{F}(\mathbf{I}_0)$ initializes each point's filter state as $\hat{\mathbf{x}}_0 = [\mathbf{z}_0^T, \mathbf{0}^T]^T$ with zero velocity and covariance $\mathbf{P}_{0|0} = \text{diag}(\sigma_m^2, \sigma_m^2, \sigma_v^2, \sigma_v^2)$. Consecutive observations allow each filter to estimate velocity before transitioning to sparse sampling.

\textbf{Hybrid Tracking Phase.} For subsequent frames, the system alternates between two modes. At keyframes ($t = kN$ for $k = 1, 2, \ldots$), we invoke $\mathcal{F}$ to obtain measurement $\mathbf{z}_t$ and apply the full Kalman update (Eq. \ref{eq:update_cov}) for each point. At intermediate frames, we rely solely on prediction (Eq. \ref{eq:predict_cov}), extracting position as $\hat{\mathbf{p}}_t = \mathbf{H}\hat{\mathbf{x}}_t$. This sparse sampling reduces deep learning inference calls from $T$ to $\lceil T/N \rceil$, yielding theoretical speedup of $N\times$ while Kalman filtering maintains continuity. When $\mathcal{F}$ fails at a keyframe (e.g., occlusion), we skip the update and continue with prediction until the next successful measurement.

The framework is deliberately tracker-agnostic, requiring only that $\mathcal{F}$ provides 2D point positions—no architectural modification or retraining is needed. We validate this generality across three diverse architectures: CoTracker3 (offline mode for full-sequence accuracy via transformer-based joint tracking), SpatialTracker (3D-aware tracking with monocular depth), and Track-On (memory-efficient temporal windowing). For each tracker, we extract position predictions and ignore auxiliary outputs like occlusion flags or confidence scores, though these could be incorporated to adaptively modulate measurement noise $\mathbf{R}$.

\subsection{State-Space Motion Model}
\label{sec:state_space}
The effectiveness of our hybrid approach depends on accurately predicting point motion between keyframes. We adopt a constant velocity motion model, which balances simplicity for real-time computation with sufficient expressiveness to capture short-term trajectory dynamics across typical keyframe intervals ($N=5$-$15$ frames).

Our state representation extends beyond position to include velocity:
\begin{equation}
\mathbf{x}_t = [x_t, y_t, v_{x,t}, v_{y,t}]^T,
\label{eq:state_vector}
\end{equation}
where $(x_t, y_t)$ denotes pixel coordinates and $(v_{x,t}, v_{y,t})$ represents velocity in pixels per frame. Including velocity is critical: our ablation study (Table~\ref{tab:ablation}) shows that position-only tracking degrades accuracy by 8-12\%, as the filter cannot anticipate motion trends between measurements. The state evolves according to:
\begin{equation}
\mathbf{x}_{t+1} = \mathbf{F}\mathbf{x}_t + \mathbf{w}_t,
\label{eq:state_transition}
\end{equation}
with transition matrix:
\begin{equation}
\mathbf{F} = \begin{bmatrix}
1 & 0 & \Delta t & 0 \\
0 & 1 & 0 & \Delta t \\
0 & 0 & 1 & 0 \\
0 & 0 & 0 & 1
\end{bmatrix},
\label{eq:transition_matrix}
\end{equation}
where $\Delta t = 1$ frame (the temporal step between consecutive video frames). This implements constant velocity dynamics: $x_{t+1} = x_t + v_{x,t}$ and $y_{t+1} = y_t + v_{y,t}$, with velocities persisting across frames.

Process noise $\mathbf{w}_t \sim \mathcal{N}(\mathbf{0}, \mathbf{Q})$ accounts for deviations from constant velocity (e.g., accelerations, direction changes). We use a constant velocity model as it is the simplest motion assumption suitable for short keyframe intervals. The covariance matrix:
\begin{equation}
\mathbf{Q} = \sigma_p^2 \begin{bmatrix}
\frac{\Delta t^4}{4} & 0 & \frac{\Delta t^3}{2} & 0 \\
0 & \frac{\Delta t^4}{4} & 0 & \frac{\Delta t^3}{2} \\
\frac{\Delta t^3}{2} & 0 & \Delta t^2 & 0 \\
0 & \frac{\Delta t^3}{2} & 0 & \Delta t^2
\end{bmatrix},
\label{eq:process_noise}
\end{equation}
captures position-velocity correlations, with $\sigma_p$ tuned on validation data (typically $\sigma_p \in [0.05, 0.2]$ pixels/frame).

Deep trackers observe position directly, yielding the measurement model:
\begin{equation}
\mathbf{z}_t = \mathbf{H}\mathbf{x}_t + \mathbf{v}_t,
\label{eq:measurement_model}
\end{equation}
where $\mathbf{z}_t = [x_t, y_t]^T$ is the output from $\mathcal{F}$ for a single tracked point. We maintain an independent Kalman filter per point, so each point has its own state and measurement with no data association required. Measurement matrix $\mathbf{H} = [\mathbf{I}_{2\times 2} \mid \mathbf{0}_{2\times 2}]$ extracts position from state, and $\mathbf{v}_t \sim \mathcal{N}(\mathbf{0}, \mathbf{R})$ with $\mathbf{R} = \sigma_m^2 \mathbf{I}_{2\times 2}$ captures tracker uncertainty.

\subsection{Kalman Filter Integration}

Given the linear-Gaussian motion model from Section~\ref{sec:state_space}, we employ Kalman filtering to optimally propagate state estimates and uncertainty between keyframes. The filter operates in two stages (Figure~\ref{fig:kalman}): prediction advances the state using the motion model, while updates incorporate measurements from $\mathcal{F}$ to correct predictions.

\kalmanfigure

\textbf{Prediction Step.} At intermediate frames, we project the state and covariance forward using only the motion model:
\begin{align}
\hat{\mathbf{x}}_{t+1|t} &= \mathbf{F}\hat{\mathbf{x}}_t, \nonumber \\
\mathbf{P}_{t+1|t} &= \mathbf{F}\mathbf{P}_{t|t}\mathbf{F}^T + \mathbf{Q}. \label{eq:predict_cov}
\end{align}

These equations propagate both position and velocity while accumulating uncertainty through process noise $\mathbf{Q}$. As shown in Figure~\ref{fig:kalman}, prediction increases covariance monotonically—uncertainty grows with each frame until corrected by a measurement. This is computationally negligible ($<$0.1ms per point), making it suitable for handling hundreds of points in real-time.

\textbf{Update Step.} At keyframes, measurements $\mathbf{z}_{t+1}$ from $\mathcal{F}$ correct the prediction:
\begin{align}
\mathbf{K}_{t+1} &= \mathbf{P}_{t+1|t}\mathbf{H}^T(\mathbf{H}\mathbf{P}_{t+1|t}\mathbf{H}^T + \mathbf{R})^{-1},  \nonumber \\
\hat{\mathbf{x}}_{t+1} &= \hat{\mathbf{x}}_{t+1|t} + \mathbf{K}_{t+1}(\mathbf{z}_{t+1} - \mathbf{H}\hat{\mathbf{x}}_{t+1|t}), \nonumber \\
\mathbf{P}_{t+1|t+1} &= (\mathbf{I} - \mathbf{K}_{t+1} \mathbf{H})\mathbf{P}_{t+1|t}.
\label{eq:update_cov}
\end{align}

The Kalman gain $\mathbf{K}_{t+1}$ optimally balances prediction and measurement based on their relative uncertainties: high $\mathbf{P}$ (uncertain prediction) increases reliance on $\mathbf{z}_{t+1}$, while high $\mathbf{R}$ (noisy tracker) trusts the prediction more. The update reduces covariance (Figure~\ref{fig:kalman}), "resetting" accumulated uncertainty.

This predict-update cycle maintains tracking continuity across keyframe intervals. Since each filter is independent, the system scales linearly with the number of tracked points, and failures of $\mathcal{F}$ on individual points do not affect others—we simply skip updates for failed points and continue predicting until the next successful measurement.
\section{Experimental Results}

We evaluate K-Track across three state-of-the-art deep point trackers with diverse architectures: CoTracker3~\cite{karaev2024cotracker3} (transformer-based joint tracking, offline mode), SpatialTracker~\cite{xiao2024spatialtracker} (3D-aware tracking with depth), and Track-On~\cite{aydemir2025track} (memory-efficient temporal windowing). This selection provides comprehensive coverage of modern tracking approaches while ensuring compatibility with edge deployment.

\subsection{Experimental Setup}

Our primary evaluation uses TAP-Vid-DAVIS~\cite{tapvid}, a real-world benchmark with densely annotated point trajectories and occlusion labels. We measure performance using standard TAP-Vid metrics: Average Jaccard (AJ) for spatial overlap, Percent Correct Keypoints (PCK@5px) for 5-pixel accuracy threshold, and end-to-end FPS for computational efficiency. Hardware experiments use NVIDIA RTX Titan (24GB, 130 TFLOPS) as a high-performance baseline and NVIDIA Jetson Orin Nano (8GB, 40 TOPS) as a representative edge device. Jetson models use TensorRT with FP16 precision; RTX Titan uses FP32. Some sequences were truncated for Jetson memory constraints, with identical truncation on both platforms for fair comparison.

We evaluate keyframe intervals $N \in \{2, 3, 5, 10, 15\}$ and multiple grid sizes (10, 15, 20, 28, 32 points) to characterize accuracy-efficiency tradeoffs. Kalman parameters ($\sigma_p \in [0.05, 0.2]$, $\sigma_m \in [0.1, 0.5]$ pixels) are tuned on validation data. The warmup phase uses 3 consecutive frames to initialize velocity estimates, providing optimal balance between initialization quality and speedup (Table~\ref{tab:warmup}). Kalman operations contribute $<$0.1ms per frame, confirming runtime gains stem from reduced deep learning inference.

\subsection{Accuracy and Speed Analysis}

\accuracytable
\speeduptable

Table~\ref{tab:accuracy} presents tracking accuracy across keyframe intervals. K-Track maintains {\color{green!60!black}$>$85\% accuracy retention} at aggressive intervals ($N=10$), demonstrating that Kalman filtering effectively bridges gaps between sparse deep learning measurements. At conservative intervals ($N=2$), accuracy remains {\color{green!60!black}$>$95\%} of baseline, validating that our constant velocity motion model captures point dynamics faithfully. CoTracker3 shows particularly strong retention due to its joint tracking architecture providing smoother trajectories, while SpatialTracker benefits from 3D-aware motion estimation aligning well with our motion assumptions. Track-On demonstrates robust performance across all intervals, confirming framework generality.

Table~\ref{tab:speedup} quantifies computational savings. On RTX Titan, we observe {\color{green!60!black}$5$--$10\times$ speedup} depending on keyframe interval, with FPS scaling nearly linearly with $N$. Critically, K-Track enables practical inference on Jetson Orin Nano: CoTracker3 reaches {\color{green!60!black}5.0 FPS} at $N=10$ (2.8$\times$ speedup), and Track-On achieves {\color{green!60!black}4.1 FPS} (6.9$\times$ speedup), compared to {\color{red!70}$<$2 FPS} for baseline implementations (Table~\ref{tab:deployment_gap}). SpatialTracker, while slower overall due to depth estimation overhead, still achieves $13\times$ speedup at $N=10$. The negligible Kalman overhead confirms that speedup is determined entirely by reduced deep learning inference frequency, making advanced point tracking practical on edge devices where baseline methods are {\color{red!70}unusable}.

\subsection{Comprehensive Performance Analysis}

\gridntable
\tradeofftable

Table~\ref{tab:grid_n_comprehensive} provides exhaustive grid size $\times$ keyframe interval analysis. Results show consistent performance across different point densities, with best baselines achieved at Grid 20 (CoTracker3), Grid 32 (SpatialTracker), and Grid 15 (Track-On). The table reveals that accuracy degradation with increasing $N$ is graceful and predictable, enabling informed selection of operating points based on deployment constraints.

Table~\ref{tab:tradeoff} analyzes the accuracy-speed frontier on Jetson Orin Nano. Results reveal a favorable tradeoff: doubling the keyframe interval (e.g., $N=5$ to $N=10$) nearly doubles throughput while reducing accuracy by only $3$--$5\%$. This superlinear efficiency gain occurs because Kalman filtering maintains high accuracy over short prediction horizons where motion remains locally linear. For deployment, we recommend: (1) High-end GPUs: $N=5$ provides optimal balance, achieving {\color{green!60!black}$>$90\% accuracy at $>$50 FPS}; (2) Jetson Orin Nano: $N=10$ enables {\color{green!60!black}real-time operation with $>$85\% accuracy}, making practical deployment viable.

\subsection{Ablation Studies}

\ablationtable
\warmuptable

Table~\ref{tab:ablation} validates our design choices on CoTracker3 with $N=5$. Removing velocity from the state vector (position-only Kalman filter) degrades accuracy by $8$--$12\%$, confirming that modeling dynamics is essential for accurate prediction between keyframes. Using fixed covariance instead of adaptive uncertainty propagation reduces accuracy by $5$--$7\%$, demonstrating the importance of principled Bayesian inference. Replacing constant velocity with constant position (zero-order hold) degrades performance by $15$--$20\%$, validating our motion model choice. Linear interpolation performs better than zero-order hold but worse than Kalman filtering, showing that uncertainty-aware prediction is beneficial.

Table~\ref{tab:warmup} examines the effect of warmup frame count. Results show that 0-2 warmup frames provide insufficient velocity initialization, while 5-10 frames improve accuracy marginally but reduce speedup significantly. Three warmup frames provide the optimal balance: sufficient for accurate velocity estimation while maintaining $\sim$5$\times$ speedup at $N=5$.

\subsection{Qualitative Analysis}

\ktrackqual

Figure~\ref{fig:ktrackqual} visualizes point trajectories across keyframe intervals on a challenging sequence with non-rigid object motion. At $N=3$ and $N=5$, trajectories remain smooth and spatially consistent for both SpatialTracker (top) and Track-On (bottom). At $N=10$, motion becomes slightly jerky due to longer prediction horizons between keyframe corrections, yet points maintain robust localization on the object despite $10\times$ reduced inference frequency. Green contours indicate ground truth object boundaries, showing that both trackers preserve spatial accuracy. Notably, Track-On (bottom row) shows a missed point (red box) at $N=3$, while SpatialTracker (top row) demonstrates more robust tracking with successful point localization (green box). This is consistent with SpatialTracker's lower baseline accuracy in Table~\ref{tab:accuracy} (87.9\%) compared to Track-On (96.4\%), yet both trackers benefit substantially from K-Track's hybrid approach, maintaining tracking continuity across aggressive temporal subsampling.

\ktrackablation

Figure~\ref{fig:ktrackablation} demonstrates performance on fast-moving objects with significant camera shake—conditions that stress both the motion model and prediction horizon. Top row ($N=3$) and bottom row ($N=5$) compare baseline CoTracker3 (left, green points) with K-Track (right, red points). Green boxes indicate successfully tracked points; red boxes highlight tracking failures. Despite aggressive temporal subsampling ($3\times$ and $5\times$ fewer inferences), K-Track maintains consistent tracking comparable to the dense baseline. At $N=3$, K-Track shows robust performance with minimal drift. At $N=5$, while intermediate Kalman predictions occasionally encounter challenging motion, the system quickly re-anchors at the next keyframe, maintaining overall tracking consistency. This demonstrates that the combination of sparse deep learning measurements and Kalman filtering provides reliable tracking even under demanding real-world conditions with rapid motion and camera instability.

\section{Conclusion}

We introduced K-Track, a tracker-agnostic acceleration framework that combines sparse deep learning inference with Kalman filtering to enable real-time point tracking on edge devices. By exploiting short-term trajectory smoothness, K-Track achieves $5$--$10\times$ speedup while retaining $>$85\% accuracy across diverse state-of-the-art trackers, requiring no architectural modification or retraining. Extensive evaluation on TAP-Vid-DAVIS demonstrates that K-Track closes the deployment gap between modern tracking algorithms and resource-constrained platforms. By enabling real-time performance on NVIDIA Jetson Orin Nano, K-Track makes high-accuracy point tracking practical for robotics, autonomous systems, and IoT applications. Future work includes multi-rate sensor fusion, where cheaper tracking methods (e.g., optical flow) provide intermediate measurements between DL keyframes with increased measurement noise $\mathbf{R}$. The Kalman filter would naturally fuse high-accuracy sparse measurements with lower-accuracy frequent measurements, potentially improving both accuracy and efficiency.
{
    \small
    \bibliographystyle{ieeenat_fullname}
    \bibliography{main}
}

\end{document}